\documentclass[11pt]{article}

\usepackage{amsmath,natbib,graphicx,amsthm}
\newcommand{\RR}{\mathbf{R}}
\newcommand{\x}{\mathbf{x}}
\newcommand{\X}{\mathcal{X}}
\newcommand{\XX}{\mathbf{R}^p}
\newcommand{\Z}{\mathcal{Z}}
\renewcommand{\P}{\mathcal{P}}
\newcommand{\z}{\mathbf{z}}

\newcommand{\ind}[1]{
  \begin{cases}
    1&\text{if~}#1\\
    0&\text{otherwise~}\end{cases}}
\newcommand{\Minimize}{\operatornamewithlimits{minimize}}
\newcommand{\argmin}{\operatornamewithlimits{argmin}}
\newcommand{\argmax}{\operatornamewithlimits{argmax}}
\newcommand{\Obj}{\text{Obj}}
\begin{document}

\title{Classification by Set Cover: The Prototype Vector Machine}
\author{Jacob Bien\footnote{Department of Statistics, \texttt{jbien@stanford.edu}}~ and Robert
  Tibshirani\footnote{Departments of Health, Research \& Policy, and
    Statistics, \texttt{tibs@stanford.edu}}\\
  \emph{Stanford University}\\
  Stanford, CA 94305, USA
}
\date{July 15, 2009}
\maketitle

\begin{abstract}
We introduce a new nearest-prototype classifier, the \emph{prototype vector
machine} (PVM).  It arises from a combinatorial optimization problem
which we cast as a variant of the set cover problem. We propose two
algorithms for approximating its solution.  The PVM selects a
relatively small number of representative points which can then be
used for classification.  It contains 1-NN as a special case.  The
method is compatible with any dissimilarity measure, making it amenable
to situations in which the data are not embedded in an underlying
feature space or in which using a non-Euclidean metric is desirable.  Indeed, we
demonstrate on the much studied ZIP code data how the PVM can reap the
benefits of a problem-specific metric.  In this example, the PVM outperforms
the highly successful 1-NN with \emph{tangent distance}, and does so
retaining fewer than half of the data points.  This
example highlights the strengths of the PVM in yielding a low-error, highly
interpretable model.  Additionally, we apply the PVM to a protein
classification problem in which a kernel-based distance is used.
\end{abstract}

\section{Introduction}
\label{sec:introduction}
Suppose we are given a set of training points
$\X=\{\x_1,\ldots,\x_n\}\subset \XX$ with corresponding class labels
$y_1,\ldots,y_n\in \{1,\ldots,L\}$ and, in addition, a set of unlabeled
points $\Z=\{\z_1,\ldots,\z_m\}\subset \XX$.  Our goal is to
choose a relatively small set of prototypes $\P_l\subseteq \Z$ for each
class $l$ in such a way that the collection $\P_1,\ldots,\P_L$
represents a summary or distillation of the training set (i.e., someone
given only $\P_1,\ldots,\P_L$ would have a good sense of
the original training data, $\X$ and $\mathbf{y}$).  While our default
choice is $\Z=\X$, we find it notationally easier to differentiate
between the two sets.  When $\Z=\X$, we are in the standard setting of
a condensation problem \citep{Ripley05}.

Having a well-selected set of prototypes $\P_1,\ldots, \P_L\subseteq \Z$
is advantageous for two main reasons: interpretability and
classification.  For domain specialists, examining a handful of
representative examples of each class can be highly informative especially
when $n$ is large (since looking through all examples from the original
data set could be overwhelming or even infeasible).  Intuitively, a
well-chosen set $\P_l\subseteq \Z$ of prototypes for class $l$ should capture the full
spread of variation within this class while also taking into account how
class $l$ differs from other classes.  Finally, the relative number of
prototypes in each class should be determined by the complexity of that class.

The other major use of the prototypes is for classification.  Once we
have prototype sets $\P_1,\ldots,\P_L$, we may classify any
new $\x\in \XX$ according to the class whose $\P_l$ contains
the nearest prototype:
\begin{align}
  \hat c(\x) = \argmin_{l}\min_{\z\in \P_l}d(\x,\z).
\end{align}
Notice that this classification rule reduces to 1-nearest-neighbors (1-NN) in the
case that $\P_l$ consists of all $\x_i\in \X$ with $y_i=l$.

In this paper, we introduce the \emph{prototype vector machine} (PVM), which
describes a particular choice for the sets $\P_1,\ldots,\P_L$.  At its
heart is the premise that $\P_l$ should consist of points that are close to many training points of class $l$ and are far from training points of
other classes.  This intuition captures the sense in which the word
``prototypical'' is commonly used.

In Section \ref{sec:PVM}, we begin with a conceptually simple optimization
criterion that describes a desirable choice for $\P_1,\ldots,\P_L$.  We
express this idea as an integer program and then in Section
\ref{sec:solving-problem} present two approximation algorithms for it.
Section \ref{sec:adapt-pvm-spec} discusses considerations for applying
the PVM most effectively to a given data set.  In Section
\ref{sec:related-work}, we give an overview of related work.  In Section
\ref{sec:exampl-simul-real} we demonstrate the PVM's effectiveness---both in terms of classification
accuracy and ease of interpretation---on a number of real data sets,
including the much-studied ZIP code
digits data set.

Finally, a note on the name:  The PVM has a
number of similarities with the Support Vector Machine: sparsity in
the samples and the slack formulation.  The PVM integer program is
an extension of the set cover problem, which we review presently.
\subsection{The set cover integer program}
Consider the two sets $\X$ and $\Z$ but without the labels $\mathbf y$.  Let $D$ be the $n\times m$ matrix of
dissimilarities, with $D_{ij}=d(\x_i,\z_j)$ for each $\x_i\in \X$ and
$\z_j\in \Z$ (note: $d$ need not be a metric), and fix $\epsilon>0$.  The goal is to find the smallest subset of
points $\P\subseteq \Z$ such that every point $\x_i\in \X$ is within
$\epsilon$ of some point in $\P$ (i.e., there exists $\z_j\in
\P$ with $d(\x_i,\z_j)<\epsilon$).  Let $B_\epsilon(\x) = \{\x'\in
\RR^p: d(\x',\x) < \epsilon \}$ denote the ball of radius $\epsilon$
centered at $\x$.  Introducing the
indicator variables 
$$\alpha_j=\begin{cases}
    1&\text{if~}\z_j \in \P\\
    0&\text{otherwise,}\end{cases}$$
this problem can be stated as an integer program:
\begin{align}\label{eq:setcover}
  \Minimize\quad&\sum_{j=1}^m \alpha_j&\nonumber\\
  \text{subject to}\quad&\sum_{j:\x_i\in B_\epsilon(\z_j)}{\alpha_j}\ge
  1&\forall~\x_i\in \X\\
  &\alpha_j \in \{0,1\}&\forall~\z_j\in \Z.\nonumber
\end{align}
The objective is simply $|\P|$.  The summation in the constraint counts the number of elements of
$\P$ that are within $\epsilon$ of the point $\x_i$, so a feasible
solution to the above integer program is one that has at least one prototype within $\epsilon$ of
each training point.

From a machine learning point of view, set cover can be seen as a
clustering problem in which we wish to find the smallest number of clusters
such that every point is within $\epsilon$ of at least one cluster
center.  In the language of vector quantization, it seeks the smallest
codebook (restricted to $\Z$) such that no vector is distorted by more
than $\epsilon$ \citep{Tipping01}.

\section{The prototype vector machine}
\label{sec:PVM}
\begin{figure}
  \centering
  \includegraphics[width=0.75\linewidth]{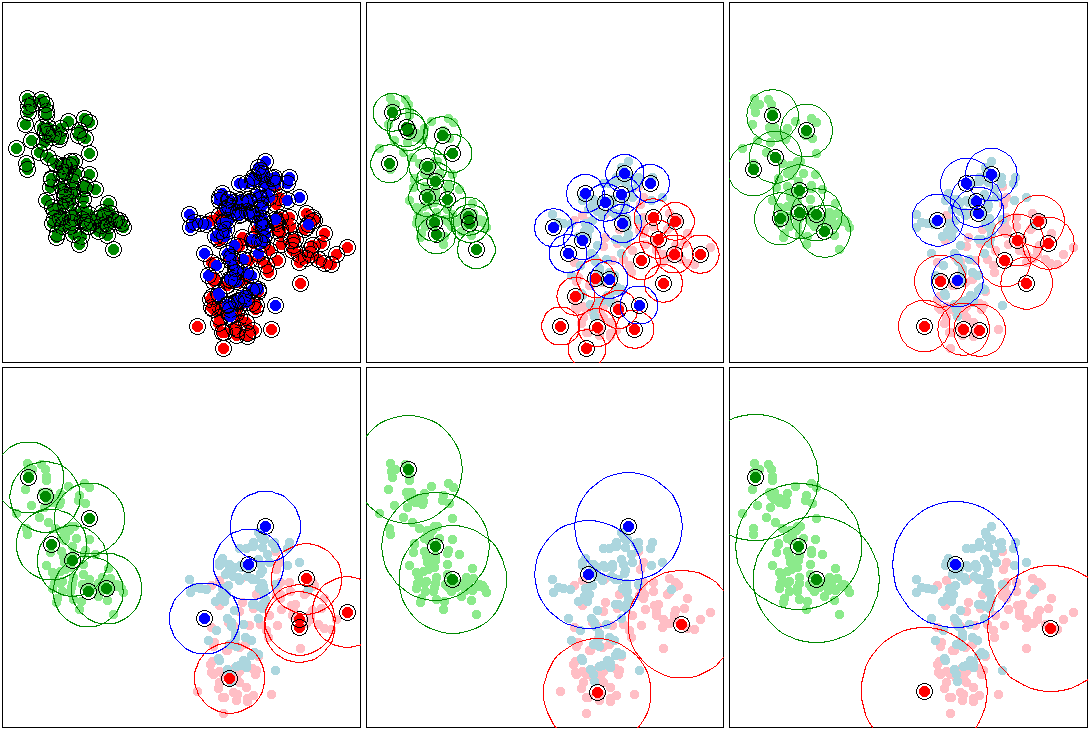}
  \caption{\em Given a value for $\epsilon$, the choice of
    $\P_1,\ldots,\P_L$ induces $L$ partial covers of the training points by
    $\epsilon$-balls centered at each prototype.  Here $\epsilon$
    is varied from the smallest interpoint distance (upper-left) to
    approximately the median interpoint distance (lower-right).}
  \label{fig:demo}
\end{figure}
The \emph{prototype vector machine} is an extension of the set
cover problem to the supervised learning context (in which each
$\x_i\in \X$ has a class label $y_i$).  The PVM seeks a set of
prototypes for each class that is optimal in a sense that will be made precise in
what follows.  For a given choice of $\P_l\subseteq \Z$, we consider the set of $\epsilon$-balls centered at
each $\z_j\in \P_l$ (see Figure \ref{fig:demo}).  A desirable prototype
set for class $l$ is one that induces a set of balls which
\begin{enumerate}
\item[(a)]\setlength{\parskip}{-3pt} covers as many training points of
  class $l$ as possible,
\item[(b)] covers as few training points as possible of classes other than $l$,
\item[and (c)] is sparse (i.e., uses as few prototypes as possible for the given $\epsilon$).
\end{enumerate}

\subsection{PVM as an integer program}
\label{sec:statement-problem}
We now express the three properties above as an integer program, taking
as a starting point the set cover problem of Equation
\ref{eq:setcover}.  Property (b) suggests that in certain cases it may
be necessary to leave some points of class $l$ uncovered.  For this reason,
we adopt a \emph{prize-collecting set cover} framework for our problem
(i.e., we assign a cost to each covering set, a penalty for being
uncovered to each point, and then find the minimum-cost partial cover,
\citealt{Konemann06}).
Let $\alpha_j^{(l)}\in \{0,1\}$ indicate whether we choose $\z_j$ to
be in $\P_l$ (i.e., to be a prototype for class $l$).  We define the
PVM to be a solution to the following integer program:
\begin{subequations}\label{eq:PVMIP}
  \begin{align}
    \Minimize_{\alpha_j^{(l)},~\xi_i,~\eta_i}\quad
    \sum_{i}{\xi_i}+&\sum_{i}{\eta_i}+\lambda\sum_{j,l}{\alpha_j^{(l)}}\nonumber\\
    \text{subject to}\quad\quad\qquad&\nonumber\\
    \sum_{j:\x_i\in B_\epsilon(\z_j)}{\alpha_j^{(y_i)}}&\ge 1-\xi_i
    \quad\forall~\x_i\in \X\label{first}\\
    \sum_{\substack{j:\x_i\in B_\epsilon(\z_j)\\l\ne y_i}}\alpha_j^{(l)}&\le
    0+\eta_i\quad\forall~\x_i\in \X\label{second}\\
    \alpha_j^{(l)}&\in \{0,1\} \quad\forall~\z_j\in \Z,l\in\{1,\ldots,L\}\nonumber\\
    \xi_i,\eta_i &\ge 0 \quad\qquad\forall~\x_i\in \X\nonumber
  \end{align}
\end{subequations}
We have introduced two slack variables, $\xi_i$ and
$\eta_i$, per training point $\x_i$.  Constraint (\ref{first})
enforces that each training point be covered
by at least one ball
of its own class-type (otherwise $\xi_i=1$).  Constraint (\ref{second})
expresses the condition that training point $\x_i$ not be covered with balls of
other classes (otherwise $\eta_i>0$).  In particular, the
slack variables can be interpreted as
\begin{itemize}\label{slack}
\item $\xi_i=\ind{\x_i \text{~is not covered by a class-$y_i$ prototype
      ball}}$
\item $\eta_i=$~Number of prototypes covering $\x_i$ that are not of class $y_i$.
\end{itemize}

Finally, $\lambda\ge0$ is a parameter specifying the cost of adding a prototype.  Its effect
is to control the number of prototypes chosen (corresponding to
property (c) of the last section).  We generally
choose $\lambda=1/n$, so that property (c) serves only as a
``tie-breaker'' for choosing among multiple solutions that
do equally well on properties (a) and (b).  Hence, in words, we are
minimizing the sum of (a) the number of points left uncovered, (b) the
number of points wrongly covered, and (c) the number of covering balls
(multiplied by $\lambda$).  The resulting method has a single tuning parameter,
$\epsilon$ (the ball radius), which can be estimated by cross validation.

We show in the Appendix that the PVM integer program is equivalent to $L$ separate prize-collecting set
cover problems.  Let $\X_l=\{\x_i\in \X:y_i=l\}$.  Then, for each
class $l$, the set $\P_l\subseteq\Z$ is given by the solution to
\begin{align*}
  \Minimize\quad\sum_{j=1}^m{C_l(j)\alpha_j^{(l)}} + \sum_{\x_i\in\X_l}{\xi_i}&\\
  \text{subject to}\quad \sum_{j:\x_i\in B_\epsilon(\z_j)}\alpha_j^{(l)}\ge 1-\xi_i
  \quad&\forall~\x_i\in \X_l\\
  \alpha_j^{(l)}\in \{0,1\} \quad&\forall~\z_j\in \Z\\
  \xi_i\ge 0 \quad&\forall~\x_i\in \X_l\\
\end{align*}
where $C_l(j)$ is the cost of adding $\z_j$ to $\P_l$ and a unit
penalty is charged for each point $\x_i$ of class $l$ left uncovered.
The cost of a covering set for the PVM is the number of miscovered
points plus a baseline charge of $\lambda$:
\begin{align*}
  C_l(j) &= \lambda + |B_\epsilon(\z_j)\cap(\X\setminus \X_l)|.
\end{align*}

\section{Solving the problem:  two approaches}
\label{sec:solving-problem}
The prize-collecting set cover problem can be transformed to a standard
set cover problem \citep{Konemann06}, which is itself NP-hard,
so we do not expect to find a polynomial-time algorithm to solve the
general PVM problem exactly.  Further, certain inapproximability
results have been proven for the set cover
problem \citep{Feige98}.\footnote{We do not assume in general that the dissimilarities
satisfy the triangle inequality, so we consider arbitrary covering
sets.}  In what follows, we present two algorithms for approximately solving our problem.
\subsection{LP relaxation with randomized rounding}
\label{sec:lp-relaxation}
A well-known approach for the set cover problem
is to relax the integer constraint $\alpha_j^{(l)}\in \{0,1\}$ by
replacing it with $0\le
\alpha_j^{(l)}\le 1$.  The result is a linear
program (LP), which is convex and easily solved with any LP solver.  The
result is subsequently rounded to recover a feasible (though not
necessarily optimal) solution to the original integer program.

Let $\{\alpha_j^{*(l)}\}$ denote a solution to the LP.  Since our
solution in general will be fractional, we adopt the following rounding strategy to produce an integral solution:  For each $j\in \{1,\ldots,m\}$ and
$l\in \{1,\ldots,L\}$, we independently draw $A_j^{(l)}\sim
\text{Bernoulli}(\alpha_j^{*(l)})$.  Notice that $\alpha_j^{*(l)}\in [0,1]$, so this approach is well-defined.  Let
$S_i$ and $T_i$ denote the slack (corresponding to $\xi_i$
and $\eta_i$) incurred by the rounded solution $\{A_j^{(l)}\}$.  These
random variables are given by 
\begin{equation}
  \label{eq:randslack}
  \begin{aligned}
    S_i &= \ind{\x_i \text{\ uncovered} \iff\sum_{j:\x_i\in B_\epsilon(\z_j)}{A_j^{(y_i)}}=0}\\
    T_i &= \sum_{l\neq y_i}\sum_{j:\x_i\in B_\epsilon(\z_j)}A_j^{(l)}.
  \end{aligned}
\end{equation}
The randomized rounding algorithm is as follows (with $B$ typically in
the hundreds):\\
\vskip .1in
\fbox{
\begin{minipage}{1.0\linewidth}
For $b=1,\ldots, B$:
\begin{enumerate}
\item Draw independently $A_j^{(l)}(b)\sim
  \text{Bernoulli}(\alpha_j^{*(l)})$.
\item Find the corresponding $S_i(b)$, $T_i(b)$ making this a feasible
  solution. (using Equation \ref{eq:randslack})
\item Evaluate objective $OBJ(b) = \sum_{i=1}^n (S_i(b)+T_i(b)) +
  \lambda\sum_{j,l}A_j^{(l)}(b)$.
\end{enumerate}
Return $\{A_j^{(l)}(b)\}$ with minimum $OBJ(b)$.
\end{minipage}
}
\vskip .1in
In the Appendix, we prove that the expected objective on
any iteration satisfies 
\begin{align*}
  E\left[\sum_{i=1}^n (S_i(b)+T_i(b)) + \lambda\sum_{j,l}A_j^{(l)}(b)\right]\
  \le \frac{n}{e} + OPT_{LP} \le \frac{n}{e} + OPT_{IP} 
\end{align*}
where $OPT_{LP} =
\sum_{i=1}^n{(\xi^*_i+\eta^*_i)}+\lambda\sum_{j=1}^n{\sum_{l=1}^L{\alpha_j^{*(l)}}}$
is the optimal value of the LP (which is a lower bound on
the integer program's optimal value).

One disadvantage of this approach is that it requires
solving an LP, which can be relatively slow and memory-intensive for
large data sets.  The approach we describe next is much lighter-weight
and is thus our preferred method.
\subsection{A greedy approach}
\label{sec:greedy-approach}
Another well-known approximation algorithm for the set cover
problem is the greedy algorithm \citep{Vazirani01}.  At each step, we add the
prototype that has the least ratio of cost to number of points
newly covered.  However, here we present a less standard greedy algorithm which has certain
practical advantages over the standard greedy approach and does not in
our experience do noticeably worse in minimizing the PVM objective.
At each step we find the $\z_j\in \Z$ and class $l$ for which adding $\z_j$ to $\P_l$ most
decreases the objective function.  That is, we find the $(\z_j,l)$ pair
with the best tradeoff of covering previously uncovered training points of
class $l$ while avoiding covering points of other classes.
The incremental improvement of going from $(\P_1,\ldots,\P_L)$ to
$(\P_1,\ldots,\P_{l-1},\P_l\cup\{\z_j\},\P_{l+1},\ldots,\P_L)$ can be
denoted by $\Delta\Obj(\z_j,l)= \Delta\xi(\z_j,l) - \Delta\eta(\z_j,l) - \lambda$ where
\begin{align*}
  \Delta\xi(\z_j,l) &=\left|\X_l\cap\left(B_\epsilon(\z_j)\setminus
      \bigcup_{\z_{j'}\in \P_l}B_\epsilon(\z_{j'})\right)\right|\\
  \Delta\eta(\z_j,l)&=|B_\epsilon(\z_j)\cap(\X\setminus \X_l)|\\
\end{align*}
The greedy algorithm is simply as follows:\\
\vskip .1in
\fbox{
\begin{minipage}{1.0\linewidth}
\begin{enumerate}
\item Start with $\P_l=\emptyset$ for each class $l$.
\item While $\Delta\Obj(\z^*,l^*)>0$:
  \begin{itemize}
  \item Find $(\z^*,l^*)=\argmax_{(\z_j,l)}{\Delta\Obj(\z_j,l)}$.
  \item Let $\P_{l^*}:= \P_{l^*}\cup\{\z^*\}$.
  \end{itemize}
\end{enumerate}
\end{minipage}
}
\section{Problem-specific considerations}
\label{sec:adapt-pvm-spec}
The PVM provides a considerable amount of flexibility that allows the
user to tailor it to the particular problem at hand.

\subsection{Dissimilarities}
\label{sec:dissimilarities}
The PVM depends on all of the $\x_i$ and $\z_j$ only through the pairwise
dissimilarities $d(\x_i,\z_j)$ and can accept any matrix with
non-negative entries.  This allows it to share in the benefits of kernel methods by using a kernel-based distance.\footnote{Given a kernel
  $K(x,x')$, we can use the distance
  $$d(x,x')=\sqrt{K(x,x)+K(x',x')-2K(x,x')}$$}  Also, for problems in the $p\gg n$
realm, using distances that effectively lower the dimension can lead
to improvements.  For instance, we have achieved gains in
classification accuracy in some $p\gg n$ simulations by using the
DANN-distance \citep{HT95}, which is a supervised measure of
distance.  Additionally, in certain problems
(e.g., in proteomics, see Section \ref{sec:prot-class-with}) the data
may not be readily embedded in a vector space.  In such a case, we may
still apply the PVM if pairwise dissimilarities are available.

Finally, given any dissimilarity $d$, we may instead use $\tilde d$,
defined by $\tilde d(\x,\z) = |\{\x_i\in \X: d(\x_i,\z) \le d(\x,\z)\}|$.  Using
$\tilde d$ induces $\epsilon$-balls $B_\epsilon(\z_j)$ containing the
$(\lfloor\epsilon\rfloor-1)$ nearest training points to $\z_j$.

\subsection{Prototypes not on training points}
\label{sec:prot-not-train}

Another inherent flexibility of the PVM is in the choice of $\Z$, the
set of potential prototypes.  While $\Z=\X$ is a standard choice, we
have experimented with other possibilities as well.  For example, if
we are also given a set of unlabeled data (e.g., a test set), we may add these
examples as potential prototypes, yielding a semi-supervised version
of the PVM.  Doing so preserves the property that all prototypes are
actual examples (rather than arbitrary points in $\XX$).

We believe that having prototypes confined to lie on actual observed
points is desirable for interpretability.  However, in circumstances
in which this property is not needed, $\Z$ may be further augmented to include other points.  For example, one could run
$K$-means on each class's points individually (or on the training set as a
whole) and add these $L\cdot K$ centroids to $\Z$.  This method seems to help
especially in high dimensional problems where constraining all
prototypes to lie on data points suffers from the \emph{curse of
dimensionality}.  Another successful choice for $\Z$ is to sample
uniformly within the convex hull of each class's training points.

\section{Related Work}
\label{sec:related-work}
Before presenting the PVM's empirical performance on data sets, we discuss
its relation to several pre-existing methods.  The PVM with $\Z=\X$
selects a subset of the original training set as prototypes.  In this
sense, it is similar in spirit to condensing and data editing methods,
such as the \emph{condensed
nearest neighbor rule} \citep{H68} and \emph{multiedit} \citep{DK82}.
\citet{H68} introduces the notion of the minimal consistent subset---the smallest subset of $\X$
for which nearest-prototype classification has 0 training error.  The
PVM objective,
$\sum_{i=1}^n{\xi_i}+\sum_{i=1}^n{\eta_i}+\lambda\sum_{j,l}{\alpha_j^{(l)}}$, represents a sort of compromise, governed by $\lambda$, between consistency
(first two terms) and minimality (third term).  In future work, we
will investigate formulations similar to PVM more closely directed toward the goal of
the minimal consistent subset.

In a similar vein, an interesting connection can be drawn to the recent work of
\citet{Weinberger09} in which they introduce \emph{large margin
  nearest neighbor classification} (LMNN), a novel approach to learning a
metric that is well-suited to $k$-NN.  LMNN seeks a linear transformation of
the feature space that brings same-class nearest neighbors closer
together and makes opposing-class
points farther apart with the goal of having each training point's $k$ nearest
neighborhood as homogenous (in class label) as possible.  The motivating
intuition is thus similar to that of the PVM, in particular properties (a) and (b) of Section \ref{sec:PVM}.  The obvious
difference between the methods is in what they output: LMNN learns a metric whereas PVM selects prototypes.

Finally, we mention a few other nearest prototype methods.
$K$-means and $K$-medoids are common unsupervised methods which produce
prototypes.  Simply running these methods on each class separately yields
prototype sets $\P_1,\ldots,\P_L$.  $K$-medoids is similar to PVM in
that its prototypes are selected from a finite set.  In contrast,
$K$-means's prototypes are not required to lie on training points,
making the method \emph{adaptive}.  Probably the most widely used
prototype method is \emph{learning vector quantization} (LVQ, \citealt{Kohonen01}).  It is an adaptive prototype method as well.
Several versions of LVQ exist, varying in certain details, but each
begins with an initial set of prototypes and then iteratively adjusts them
in a fashion that tends to encourage each prototype to lie near many training points of its class and away from training
points of other classes.

\section{Examples on simulated and real data}
\label{sec:exampl-simul-real}
We compare the PVM's perfomance to some of the prototype methods
mentioned above.  For
$K$-medoids, we run \texttt{pam} of the R package \texttt{cluster} on
each class's data separately, producing $K$ prototypes per class.

For LVQ, we use the functions \texttt{lvqinit} and \texttt{olvq1} (optimized
learning vector quantization 1, \citealt{Kohonen01}) from the R
package \texttt{class}.  We vary the initial codebook size to produce a range of
solutions.

\subsection{Mixture of Gaussians simulation}
\label{sec:mixt-gauss-simul}
\begin{figure}
  \centering
  \includegraphics[width=1.0\linewidth]{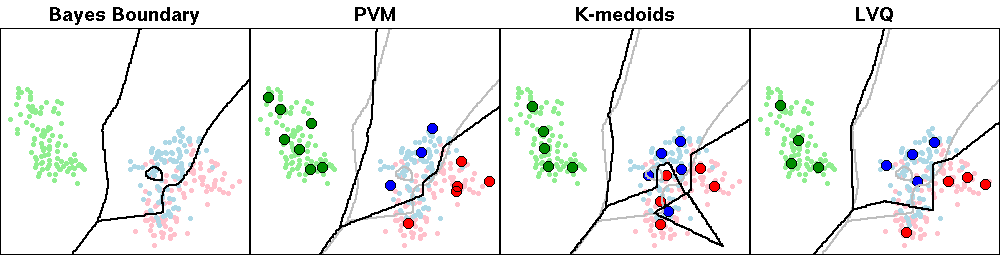}
  \caption{\em Mixture of Gaussians training data.  Classification
    boundaries of (left to right) Bayes, PVM-Greedy, $K$-medoids,
    LVQ (with Bayes boundary shown in gray for comparison).}
  \label{fig:mixture}
\end{figure}

For demonstration purposes, we consider a three-class example with
$p=2$.  Each class was generated as a mixture of 10 Gaussians (details given in the Appendix).  Figure \ref{fig:demo} shows the PVM solution
for a range of values of the tuning parameter $\epsilon$.  In Figure
\ref{fig:mixture}, we display the classification boundaries for the PVM, $K$-medoids, and LVQ (taking the lowest test error
solution for each method).  Since we generated this example from a
known model, we are able to compute the Bayes boundary.  We see that
the PVM succeeds in capturing the shape of the boundary.  The erratic boundary of
$K$-medoids highlights an advantage of the PVM over $K$-medoids;  the
latter does not consider the relation between classes when choosing
prototypes and therefore does not perform well when classes overlap.

\subsection{ZIP code digits data}
\label{sec:digits-data}
We apply the PVM to the USPS handwritten digits data set which consists of
a training set of $n=7291$ grayscale ($16\times 16$ pixel) images of
handwritten digits 0-9 (and $2007$ test images).  We run the PVM for a range
of values of $\epsilon$ from the minimum interpoint distance (in which
the PVM retains the entire training set and so reduces to 1-NN classification)
to approximately the $14^{\text{th}}$ percentile of interpoint
distances.

\begin{figure}
  \centering
  \includegraphics[width=0.45\linewidth]{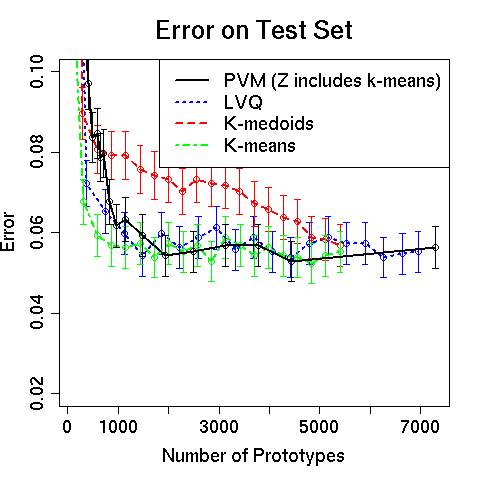}
  \includegraphics[width=0.45\linewidth]{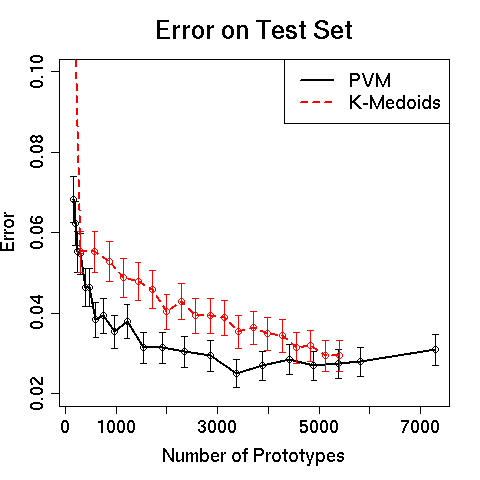}
  \caption{\em Digits data set.  (Left) All methods use Euclidean distance (Right)
    Both use tangent distance.  The rightmost point on the PVM
    curves correspond to 1-NN classification.}
  \label{fig:digitstest}
\end{figure}

The lefthand panel of Figure \ref{fig:digitstest} shows the test error
as a function of the number of prototypes for several methods
using the Euclidean metric.  Since both LVQ and $K$-means can place
prototypes anywhere in the feature space, which is advantageous in
high-dimensional problems, we also allow PVM to select prototypes that do
not lie on the training points by augmenting $\Z$.  In this case, we
run 10-means clustering on each class separately and then add these
resulting 100 points to $\Z$ (in addition to $\X$).

The notion of the \emph{tangent distance} between two such images was introduced by \citet{sim-93} to account for certain invariances
in this problem (e.g., the thickness and orientation of a digit
are not relevant factors when we consider how similar two digits
are).  Use of tangent distance with 1-NN attained the lowest test
errors of any method \citep{HS97}.  Since the PVM operates on
an arbitrary dissimilarities matrix, we can easily use the tangent distance
in place of the standard Euclidean metric.
The righthand panel of Figure \ref{fig:digitstest} shows the test
errors when tangent distance is used.  $K$-medoids similarly
readily accommodates any dissimilarity.  While LVQ has been generalized to
arbitrary differentiable metrics, there does not appear to be
generic, off-the-shelf software available.  The lowest test error attained
by the PVM is 2.49\% with a 3372-prototype solution (compared to 1-NN's
3.09\%).\footnote{\citet{HS97} report a 2.6\% test error for 1-NN
  on this data set.  The difference may be due to implementation details
  of the tangent distance.}  Also, we can see that for a
wide range of $\epsilon$ values we get a solution with test
error comparable to that of 1-NN, but requiring far fewer prototypes.
An advantageous feature of the PVM is that it automatically chooses
the number of prototypes per class to use.  In this example, it is interesting to see the class-frequencies
of prototypes (see Table \ref{tab:numprotos}).

\begin{table}[h]
\centering
\small
\begin{tabular}{r|cccccccccc|c}
  \bf Digit& 0&1&2&3&4&5&6&7&8&9&\bf Total\\
  \hline
  \bf Training set&1194& 1005&  731&  658&  652&  556&  664&  645&  542&
  644& 7291\\
  \bf PVM-best& 493&   7& 661& 551& 324& 486& 217& 101& 378& 154& 3372\\
\end{tabular}
\caption{\em Number of prototypes chosen per class}
\label{tab:numprotos}
\end{table}

The most dramatic feature of this solution is that it only retains seven
of the 1005 examples of the digit 1.  This reflects the fact that,
relative to other digits, the digit 1 has the least variation when
handwritten.  Indeed, the average (tangent) distance between digit 1's in the
training set is less than half that of any other digit (the second
least variable digit is 7).

\begin{figure}
  \centering
  \includegraphics[width=1.0\linewidth]{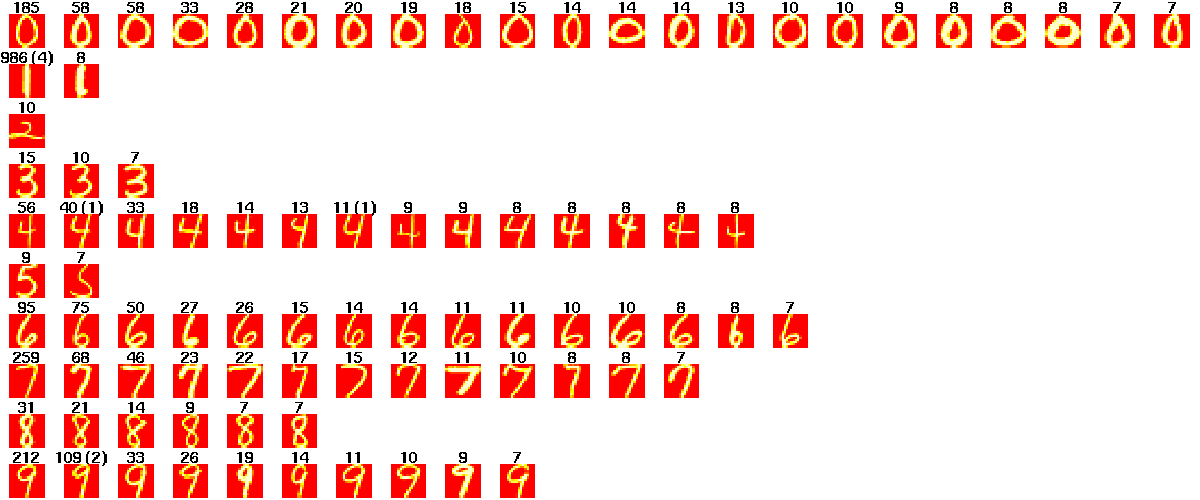}
  \caption{\em First 88 (of 3372) PVM-Greedy prototypes.  Above each is
    the number of training images first correctly covered by the addition of this prototype (in parentheses is the
    number of miscovered training points by this prototype).}
  \label{fig:digits_protos}
\end{figure}

\begin{figure}
  \centering
  \includegraphics[width=0.8\linewidth]{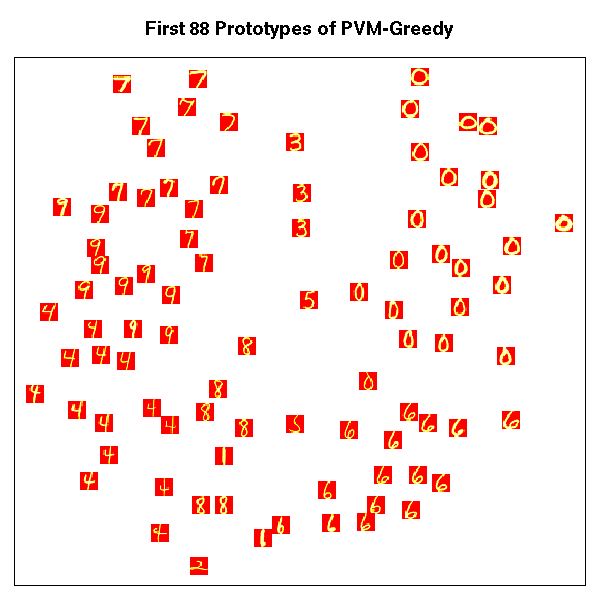}
  \caption{\em The first 88 prototypes (out of 3372) of the PVM-Greedy
    solution.  We perform MDS (\texttt{sammon}, stress=0.07) on the
    tangent distances to visualize the prototypes in two dimensions.}
  \label{fig:88proto_mds}
\end{figure}

\begin{figure}
  \centering
  \includegraphics[width=0.6\linewidth]{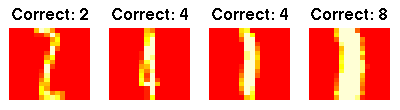}
  \caption{\em The four training images that were miscovered by the first
    prototype of class 1 (see Figure \ref{fig:digitstest}).}
  \label{fig:miscovered}
\end{figure}

In this example, we took $\Z=\X$, so that each prototype is an actual
handwritten digit from the training set (rather than being some linear
combination of many handwritten digits).  Figures
\ref{fig:digits_protos} and \ref{fig:88proto_mds} show
images of the first 88 prototypes (of 3372) selected by the greedy
algorithm.  Above each image of Figure \ref{fig:digits_protos} is the number of training images
previously uncovered that were correctly covered by the addition of
this prototype and, in parentheses, the number of training points that are miscovered by this prototype.  For example, we can
see that the first prototype selected by the greedy algorithm, which was a
``1,'' covered 986 training images of 1's and four training
images that were not of 1's.  These four training images are shown in
Figure \ref{fig:miscovered}.  Indeed, all of them look very much like
1's, which explains the algorithm's confusion.

\begin{figure}
  \centering
  \includegraphics[width=0.45\linewidth]{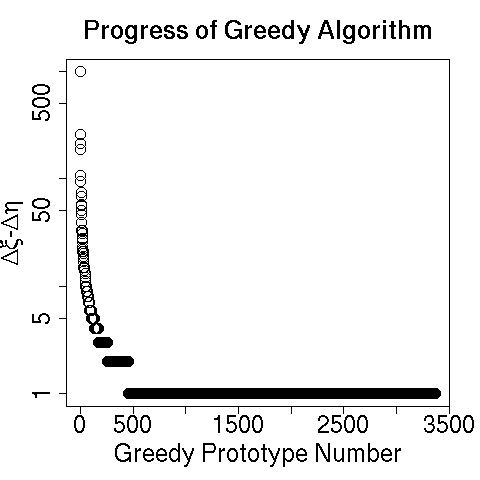}
  \includegraphics[width=0.45\linewidth]{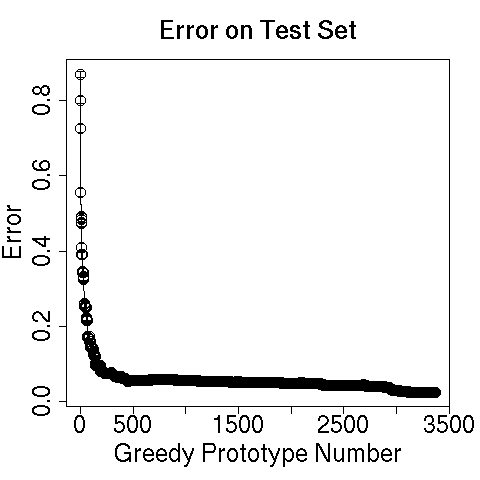}
  \caption{\em Progress of greedy as a function of number of protoypes added.}
  \label{fig:greedyscore}
\end{figure}

The lefthand panel of Figure \ref{fig:greedyscore} shows the improvement in the PVM
objective, $\Delta\xi - \Delta\eta$, after each step of the greedy
algorithm, revealing an interesting feature of the solution: we find
that after the first 458 prototypes are added, each remaining
prototype covers only one training point.  Since in this example we
took $\Z=\X$ (and since a point always covers itself), this means that the final 2914 prototypes were chosen
to cover only themselves.  In this sense, we see that the PVM provides
a sort of compromise between a sparse nearest prototype classifier and
1-NN.  The compromise is determined by the
prototype-cost parameter $\lambda$.  If $\lambda>1$, the algorithm
does not enter the 1-NN regime.

\begin{figure}
  \centering
  \includegraphics[width=.75\linewidth]{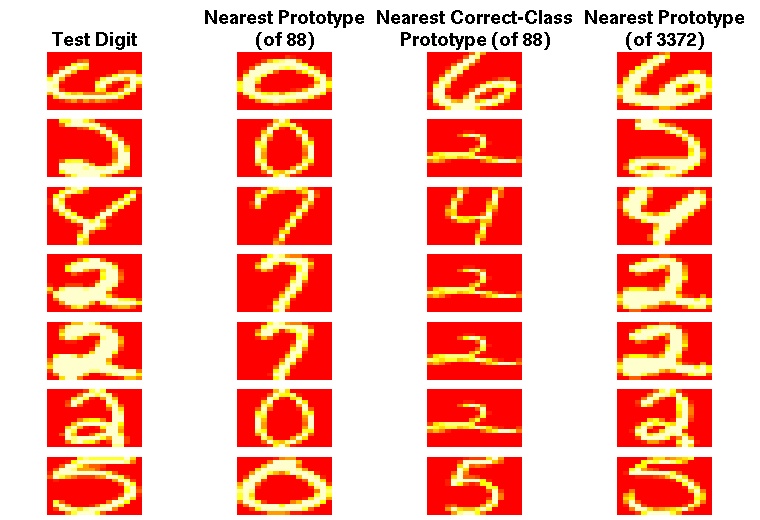}
  \caption{\em Each row corresponds to a test digit that is misclassified using just the first 88 prototypes (which are shown in Figure \ref{fig:digits_protos}).  From left to right: the test digit itself, the nearest prototype among the 88 prototypes, the nearest prototype of the correct class (among the 88), and the nearest prototype in the full 3372-prototype solution.}
  \label{fig:corrected}
\end{figure}

The righthand panel of Figure \ref{fig:greedyscore} shows the
improvement in test error gained by running the greedy algorithm
beyond the first 88 steps (corresponding to the $\lambda=6$
solution).  It is interesting to look at elements of the test set that
are misclassified when we use just the 88 prototypes but are correctly
classified when using the complete PVM-greedy solution (with all 3372 prototypes).  There are 276 (out of
2007) such elements.  Figure \ref{fig:corrected} shows a randomly chosen seven
examples of these test points.

\subsection{Protein Classification with String Kernels}
\label{sec:prot-class-with}

\begin{figure}
  \centering
  \includegraphics[width=0.5\linewidth]{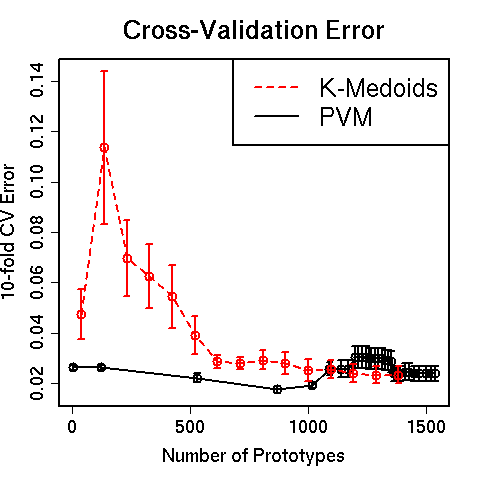}
  \caption{\em Proteins data set.  Recall that the rightmost point on the PVM
    curve corresponds to 1-NN classification.}
  \label{fig:stringkerneltest}
\end{figure}

In our next example, we present a case in which the patterns are not
naturally represented as vectors in $\RR^p$.  \citet{Leslie04} study the problem of classification of proteins based on their
amino acid sequences.  They introduce a measure of similarity between
protein sequences called the \emph{mismatch kernel}.  The general idea is that two
sequences should be considered similar if they have a large number of
short sequences in common (where two short sequences are considered
the same if they have no more than a specified number of mismatches).
We take as input a $1708\times 1708$ matrix with $K_{ij}$ containing
the value of the normalized mismatch kernel evaluated between proteins
$i$ and $j$ (the data and software are from \citealt{Leslie04}).  The proteins fall into two classes, ``Positive'' and
``Negative,'' according to whether they belong to a certain protein
family.  We compute pairwise distances from this kernel via $D_{ij} =
\sqrt{K_{ii}+K_{jj}-2K_{ij}}$ and then run the PVM and $K$-medoids.  Figure \ref{fig:stringkerneltest}
shows the 10-fold cross-validated errors for the PVM and $K$-medoids.  For
the PVM, we take a range of equally-spaced quantiles of the pairwise
distances from the minimum to the median for the parameter
$\epsilon$.   For $K$-medoids, we take as parameter the fraction of
proteins in each class that should be prototypes.  This choice of
parameter allows the classes to have different numbers of prototypes,
which is important in this example because the classes are
greatly imbalanced (only 45 of the 1708 proteins are in class
``Positive'').  The minimum CV-error (1.76\%) is attained by PVM using
about 870 prototypes (averaged over the 10 models fit for that value
of $\epsilon$).  This error is identical to the minimum CV-error of a
support vector machine (tuning the cost parameter) trained using this kernel.
Fitting a model to the whole data set with the selected value of $\epsilon$, the PVM chooses 26 prototypes (of 45) for class ``Positive'' and 907 (of
1663) for class ``Negative.''

\subsection{UCI data sets}
\label{sec:uci-data-sets}
Finally, we run the PVM on six data sets from the UCI Machine Learning
Repository \citep{UCI} and compare its performance to that of 1-NN
(i.e., retaining all training points as prototypes), $K$-medoids,
and LVQ.  We randomly select 2/3 of each data set for training and use
the remainder as a test set.  Ten-fold cross-validation (and the
``1 standard error
rule,'' \citealt{ESL}) is performed on the training data to select a
value for each method's tuning parameter (except for 1-NN).  Table
\ref{tab:uci} reports the error on the test set and the number of
prototypes selected for each method.  We see that in most cases PVM is able to do as
well as or better than 1-NN but with a significant reduction in
prototypes.  No single method does best on all of the data sets.
\begin{table}
  \centering
  \begin{tabular}{lr|c|c|c|c|}
    \textbf{Data}&&\textbf{1-NN}&\textbf{PVM}&\textbf{$K$-medoids}&\textbf{LVQ}\\
    \hline
    Diabetes&\footnotesize{\emph{Test Error (\%)}}&28.9&24.2&33.2&25.0\\
    \footnotesize{($p=8,L=2$)}&\footnotesize{\emph{\# Prototypes}}&512&12&44&29\\
    \hline
    Glass&\footnotesize{\emph{Test Error (\%)}}&38.0&36.6&39.4&35.2\\
    \footnotesize{($p=9,L=6$)}&\footnotesize{\emph{\# Prototypes}}&143&34&12&17\\
    \hline
    Heart&\footnotesize{\emph{Test Error (\%)}}&21.1&21.1&17.8&15.6\\
    \footnotesize{($p=13,L=2$)}&\footnotesize{\emph{\# Prototypes}}&180&6&26&12\\
    \hline
    Liver&\footnotesize{\emph{Test Error (\%)}}&41.7&41.7&40.0&33.9\\
    \footnotesize{($p=6,L=2$)}&\footnotesize{\emph{\# Prototypes}}&230&16&20&110\\
    \hline
    Vowel&\footnotesize{\emph{Test Error (\%)}}&2.8&2.8&2.8&19.9\\
    \footnotesize{($p=10,L=11$)}&\footnotesize{\emph{\# Prototypes}}&352&352&198&193\\
    \hline
    Wine&\footnotesize{\emph{Test Error (\%)}}&3.4&11.9&6.8&3.4\\
    \footnotesize{($p=13,L=3$)}&\footnotesize{\emph{\# Prototypes}}&119&4&12&3\\
    \hline
  \end{tabular}
  \caption{\em Test errors for the UCI data sets.  For PVM, $K$-medoids,
    and LVQ, we used 10-fold cross validation (with the 1 SE rule) on
    the training set to tune the parameters.}
  \label{tab:uci}
\end{table}

\section{Discussion}
We have introduced a new prototype method, which can be used both for
classification and for ``summarizing'' a data set.  The PVM is the solution
to a set cover problem which describes our notion of a desirable
prototype set.  Applying the PVM to the digits data highlights some of its
strengths.  First, it has competitive test error for a wide range of values of the
tuning parameter.  Its success in this example stems in part from its
flexibility: it was easily used with a problem-specific measure
of dissimilarity.  Additionally, it automatically chooses a suitable
number of prototypes for each class.  Particularly useful for
interpretation is the fact that each PVM-prototype is an observation in the
training set (i.e., is an actual hand drawn image).  In medical
applications, this would mean that prototypes correspond to actual
patients.  This feature may be of great practical use to domain
experts for making sense of large data sets.

The PVM software will be made available as an R package in the R library.

\section{Acknowledgements}
\label{sec:acknowledgements}
We thank Sam Roweis for pointing us to set cover as a clustering
method, Sam Roweis and Amin Saberi for helpful discussions, and
Trevor Hastie for providing us with his code for computing tangent
distance.  Jacob Bien is supported by the Urbanek Family Stanford
Graduate Fellowship and Robert Tibshirani was partially supported by National Science Foundation Grant DMS-9971405 and National Institutes of Health Contract N01-HV-28183.

\vskip 0.2in
\bibliography{/home/jbien/private/tibs}

\appendix
\section{PVM's relation to prize-collecting set cover}
\label{sec:pvm-integer-program}
\textbf{Claim:~}\emph{Solving the PVM integer program is equivalent
  to solving $L$ prize-collecting set cover problems.}
\begin{proof}
  Recall that the PVM integer program is given by
  \begin{align*}
    \Minimize_{\alpha_j^{(l)},~\xi_i,~\eta_i}\quad
    \sum_{i=1}^n{\xi_i}+&\sum_{i=1}^n{\eta_i}+\lambda\sum_{j,l}{\alpha_j^{(l)}}\\
    \text{subject to}\quad\quad\qquad&\\
    \sum_{j:\x_i\in B_\epsilon(\z_j)}{\alpha_j^{(y_i)}}&\ge 1-\xi_i
    \quad\forall~\x_i\in \X\\
    \sum_{\substack{j:\x_i\in B_\epsilon(\z_j)\\l\ne y_i}}\alpha_j^{(l)}&\le
    0+\eta_i\quad\forall~\x_i\in \X\\
    \alpha_j^{(l)}&\in \{0,1\} \quad\forall~\z_j\in \Z,l\in\{1,\ldots,L\}\\
    \xi_i,\eta_i &\ge 0 \quad\qquad\forall~\x_i\in \X.
  \end{align*}
  Now, the second set of inequality constraints is always tight, so we can
  eliminate the slack variables $\eta_1,\ldots,\eta_n$:
  \begin{align*}
    \Minimize_{\alpha_j^{(l)},~\xi_i}\quad
    \sum_{i=1}^n{\xi_i}+&\sum_{i=1}^n{\sum_{\substack{j:\x_i\in B_\epsilon(\z_j)\\l\ne y_i}}\alpha_j^{(l)}}+\lambda\sum_{j,l}{\alpha_j^{(l)}}\\
    \text{subject to}\quad\quad\qquad&\\
    \sum_{j:\x_i\in B_\epsilon(\z_j)}{\alpha_j^{(y_i)}}&\ge 1-\xi_i
    \quad\forall~\x_i\in \X\\
    \alpha_j^{(l)}&\in \{0,1\} \quad\forall~\z_j\in \Z,l\in\{1,\ldots,L\}\\
    \xi_i &\ge 0 \quad\qquad\forall~\x_i\in \X.
  \end{align*}
  We can rewrite the second term
  of the objective as
  \begin{align*}
    \sum_{i=1}^n{\sum_{\substack{j:\x_i\in B_\epsilon(\z_j)\\l\ne
          y_i}}\alpha_j^{(l)}} &=
    \sum_{i=1}^n{\sum_{j,l}{1\{\x_i\in B_\epsilon(\z_j),l\ne y_i\}}\alpha_j^{(l)}}\\
    &= \sum_{j,l}\alpha_j^{(l)}\sum_{i=1}^n 1\{\x_i \in
    D_j(\epsilon),\x_i\notin \X_l\}\\
    &= \sum_{j,l}\alpha_j^{(l)}|B_\epsilon(\z_j)\cap(\X\setminus \X_l)|
  \end{align*}
  So the entire objective becomes
  \begin{align*}
    \sum_{i=1}^n{\xi_i} + \sum_{j,l}\left[(|B_\epsilon(\z_j)\cap(\X\setminus \X_l)|+\lambda)\alpha_j^{(l)}\right]
  \end{align*}
  Letting $C_l(j) = \lambda + |B_\epsilon(\z_j)\cap(\X\setminus \X_l)|$,
  the integer program may be written as
  \begin{align*}
    \Minimize_{\alpha_j^{(l)},~\xi_i}\quad
    \sum_{l=1}^L&{\left[\sum_{\x_i\in\X_l}{\xi_i}+\sum_{j=1}^m{ C_l(j)\alpha_j^{(l)}}\right]}\\
    \text{subject to},~\forall~l&\in \{1,\ldots,L\},\\
    \sum_{j:\x_i\in B_\epsilon(\z_j)}{\alpha_j^{(l)}}&\ge 1-\xi_i
    \quad\forall~\x_i\in \X_l\\
    \alpha_j^{(l)}&\in \{0,1\} \quad\forall~\z_j\in \Z\\
    \xi_i &\ge 0 \quad\qquad\forall~\x_i\in \X_l.
  \end{align*}
  Written in this way, we see that both the objective and the
  constraints are separable with respect to class, meaning that the
  solution to the above is equivalent to solving $L$ integer programs
  (one for each $l\in \{1,\ldots,L\}$).  The $l^{\text{th}}$ integer program has variables $\alpha_1^{(l)},\ldots,\alpha_m^{(l)}$ and
  $\{\xi_i:\x_i\in \X_l\}$ and is given by
  \begin{align*}
    \Minimize\quad\sum_{j=1}^m{C_l(j)\alpha_j^{(l)}} + &\sum_{\x_i\in\X_l}{\xi_i}\\
    \text{subject to}\quad \sum_{j:\x_i\in B_\epsilon(\z_j)}\alpha_j^{(l)}\ge 1-\xi_i
    \quad&\forall~\x_i\in \X_l\\
    \alpha_j^{(l)}\in \{0,1\} \quad&\forall~\z_j\in \Z\\
    \xi_i\ge 0 \quad&\forall~\x_i\in \X_l.
  \end{align*}
  This is precisely the prize-collecting set cover problem (with unit
  penalty for leaving a point uncovered).
\end{proof}
\section{Randomized Rounding Bound}
\label{sec:rand-round-result}
\textbf{Claim:} \emph{Given the randomized rounding procedure
  described in Section \ref{sec:lp-relaxation}, the objective on each iteration satisfies}
\begin{align*}
  E[OBJ] \le \frac{n}{e} + OPT_{IP}.
\end{align*}
\begin{proof}
  Let $\{\alpha_j^{*(l)},\xi_i^*,\eta_i^*\}$ denote a solution to the LP and
  recall that for each iteration,  we sample independently
  \begin{align*}
    A_j^{(l)}\sim\text{Bernoulli}(\alpha_j^{*(l)}).
  \end{align*}
  The PVM objective on this iteration is given by
  \begin{align*}
    OBJ = \sum_{i=1}^n (S_i+T_i) + \lambda\sum_{j,l}A_j^{(l)}
  \end{align*}
  where
  \begin{align*}
    S_i &= \ind{\x_i \text{\ uncovered} \iff\sum_{j:\x_i\in B_\epsilon(\z_j)}{A_j^{(y_i)}}=0}\\
    T_i &= \sum_{l\neq y_i}\sum_{j:\x_i\in B_\epsilon(\z_j)}A_j^{(l)}
  \end{align*}
  Now, by linearity of expectation, we have
  \begin{align*}
    E[OBJ] = E\left[\sum_{i=1}^n (S_i+T_i) +
      \lambda\sum_{j,l}A_j^{(l)}\right] = \sum_{i=1}^n (P[\x_i \text{\ uncovered}]+\eta_i^*) + \lambda\sum_{j,l}\alpha_j^{*(l)}\\
  \end{align*}
  since $E[T_i] = \sum_{l\neq y_i}\sum_{j:\x_i\in B_\epsilon(\z_j)}\alpha_j^{*(l)} = \eta_i^*$.
  
  Now,
  \begin{align*}
    P(\x_i~\text{uncovered}) &= P\left(A_j^{(y_i)}=0~\forall~j:\x_i\in B_\epsilon(\z_j)\right)\\
    &= \prod_{j:\x_i\in B_\epsilon(\z_j)}\left( 1 - \alpha_j^{*(y_i)} \right)\\
    &\le e^{-\sum_{j:\x_i\in B_\epsilon(\z_j)}\alpha_j^{*(y_i)}}\\
    &\le e^{-(1-\xi_i^*)}\\
  \end{align*}
  using that $1-x\le e^{-x}$ and, by LP feasibility (Constraint 3a), that $-\sum_{j:\x_i\in B_\epsilon(\z_j)}\alpha_j^{*(y_i)}\le -(1-\xi_i^*)$.  Now, $0\le\xi_i^*\le1$ and
  \begin{align*}
    e^{x-1}\le \frac{1}{e} + \frac{e-1}{e}x \qquad\text{for}\qquad 0\le x\le1
  \end{align*}
  so
  \begin{align*}
    P(\x_i~\text{uncovered})\le \frac{1}{e} + \frac{e-1}{e}\xi_i^*
  \end{align*}
  from which it follows that
  \begin{align*}
    E[OBJ]&\le \sum_{i=1}^n (\frac{1}{e} +
    \frac{e-1}{e}\xi_i^*+\eta_i^*) + \lambda\sum_{j,l}\alpha_j^{*(l)}\\
    &\le \frac{n}{e} + \sum_{i=1}^n(\xi_i^*+\eta_i^*) +
    \lambda\sum_{j,l}\alpha_j^{*(l)}\\
    &= \frac{n}{e} + OPT_{LP}\\
    &\le \frac{n}{e} + OPT_{IP}\\
  \end{align*}
\end{proof}

\section{Mixture of Gaussians example}
\label{sec:mixt-gauss-example}

We generate the data of Section \ref{sec:mixt-gauss-simul} in the
style of \citet{ESL}, Section 2.3.3.  In particular,

\begin{itemize}
\item Fix 3 class centers $M_1,M_2,M_3\in\mathbf{R}^2$, sampled from $N(0,16I_2)$.
\item For each class $k$, independently generate
  $m_1^{(k)},\ldots,m_{10}^{(k)}\sim N(M_k,I_2)$.
\item For $i=1,\ldots,n$, choose $j\in
  \{1,\ldots,10\}$ uniformly at random, then draw
\begin{align*}
  \x_i|y_i \sim ~ N(m_j^{(y_i)},I_2/5).
\end{align*}
\end{itemize}

We take $n=300$ in this case, with 100 points in each class.

\end{document}